\documentclass[letterpaper]{article} 
\usepackage{aaai20}  
\usepackage{times}  
\usepackage{helvet} 
\usepackage{courier}  
\usepackage[hyphens]{url}  
\usepackage{graphicx} 
\usepackage{graphicx}  
\usepackage{multirow}
\usepackage{booktabs}
\usepackage{mathtools}
\usepackage{amsfonts}

\usepackage{float}

\usepackage[usenames,dvipsnames,svgnames,table]{xcolor}

\providecommand{\task}{NAME}

\title{Automatic Generation of Headlines for Online Math Questions}
\author{ \Large \textbf{Ke Yuan\textsuperscript{\rm 1 \rm 2}, Dafang He\textsuperscript{\rm 2}, Zhuoren Jiang\textsuperscript{\rm 3}, Liangcai Gao\textsuperscript{\rm 1}\thanks{are the corresponding authors}, Zhi Tang\textsuperscript{\rm 1}$^*$, C. Lee Giles\textsuperscript{\rm 2}} \\ 
\textsuperscript{\rm 1}Wangxuan Institute of Computer Technology, Peking University, Beijing, 100080, China\\ 
\textsuperscript{\rm 2} The Pennsylvania State University, University Park, PA 16802, USA\\
\textsuperscript{\rm 3} School of Data and Computer Science, Sun Yat-sen University, Guangzhou, 510006, China\\
\textsuperscript{}
yuanke@pku.edu.cn, duh188@psu.edu, jiangzhr3@mail.sysu.edu.cn, \{glc, tangzhi\}@pku.edu.cn, giles@ist.psu.edu
}
 
\begin{document}

\maketitle

\begin{abstract}
Mathematical equations are an important part of dissemination and communication of scientific information. Students, however, often feel challenged in reading and understanding math content and equations. With the development of the Web, students are posting their math questions online. Nevertheless, constructing a concise math headline that gives a good description of the posted detailed math question is nontrivial. In this study, we explore a novel summarization task denoted as ge\textbf{N}erating \textbf{A} concise \textbf{M}ath h\textbf{E}adline from a detailed math question (\textbf{\task}). Compared to conventional summarization tasks, this task has two extra and essential constraints: 1) Detailed math questions consist of text and math equations which require a unified framework to jointly model textual and mathematical information; 2) Unlike text, math equations contain semantic and structural features, and both of them should be captured together. To address these issues, we propose MathSum, a novel summarization model which utilizes a pointer mechanism combined with a multi-head attention mechanism for mathematical representation augmentation. The pointer mechanism can either copy textual tokens or math tokens from source questions in order to generate math headlines. The multi-head attention mechanism is designed to enrich the representation of math equations by modeling and integrating both its semantic and structural features.
For evaluation, we collect and make available two sets of real-world detailed math questions along with human-written math headlines, namely EXEQ-300k and OFEQ-10k. Experimental results demonstrate that our model (MathSum) significantly outperforms state-of-the-art models for both the EXEQ-300k and OFEQ-10k datasets.
\end{abstract}

\section{Introduction}
\begin{figure}
    \centering
    \includegraphics[scale=0.7]{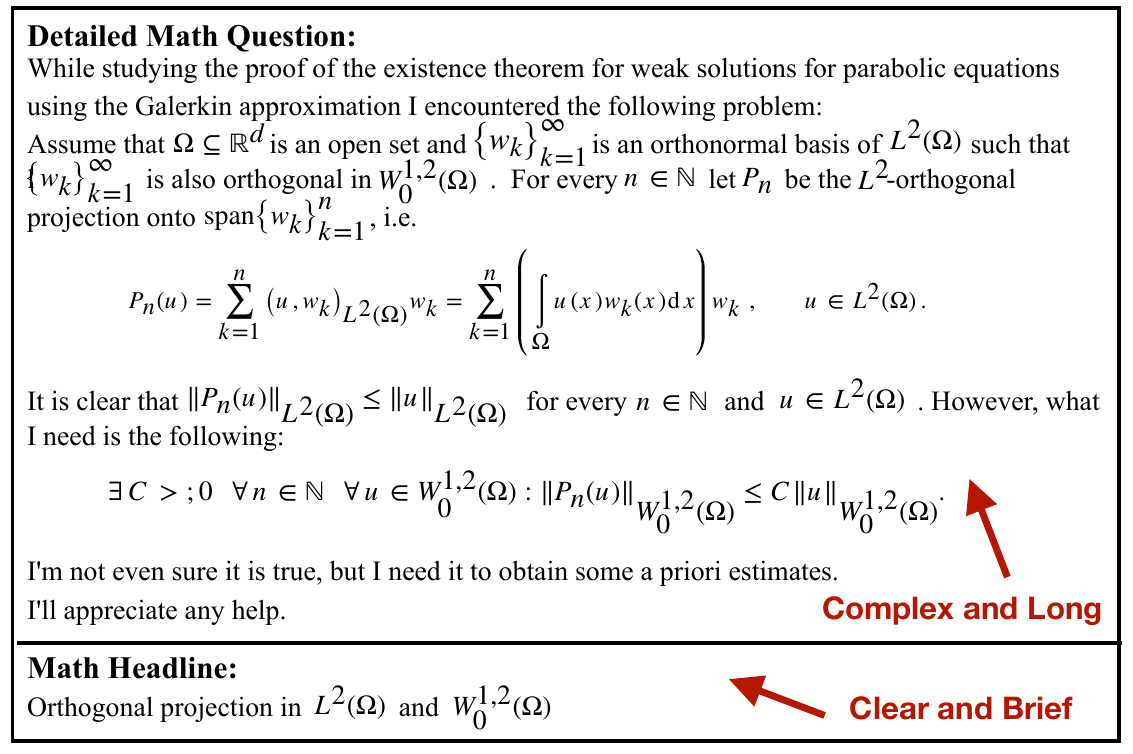}
    \caption{Example of a detailed math question along with its headline. The question is complex and long and the headline is clear and brief.}
    \label{fig:inex}
\end{figure}

Math equations are widely used in the fields of Science, Technology, Engineering, and Mathematics (STEM). However, it is often daunting for students to understand math content and equations when they are reading STEM publications~\cite{liu2014interactive,jiang2018mathematics}. Because of the Web, students post detailed math questions online for help. Recent question systems, such as \textit{Mathematics Stack Exchange\footnote{https://math.stackexchange.com}} and \textit{MathOverflow\footnote{https://mathoverflow.net}}, attempt to address this need. From the viewpoint of  questioners, the contents of detailed math questions are usually complex and long. In order to efficiently help those who pose the question, it would be helpful to have a headline which is concise and to the point. Correspondingly, those who will answer the question (answerers) also need a clear and brief headline to quickly determine if they should bother to respond. Therefore, giving a concise math headline to a detailed question is important and meaningful. Figure~\ref{fig:inex} illustrates an example of the question along with its headline posted in \textit{Mathematics Stack Exchange}\footnote{https://math.stackexchange.com/questions/3331385}. It's clear that, a complicated question can make it difficult for answerers to understand the intent of the questioner, while a concise headline can effectively reduce the cost of this operation.


To this end, we explore a novel approach for ge\textbf{N}erating \textbf{A} \textbf{M}ath h\textbf{E}adline for detailed questions (\textbf{\task}). Here, we define the {\task} task as a summarization task. Compared to conventional summarization tasks, the {\task} task has two extra essential issues that need to be addressed: 1) Jointly modeling text and math equations in a unified framework. Textual (words) and mathematical (equations) information are usually coexisting in detailed questions and brief headlines, as shown in Figure~\ref{fig:inex}. As such, it is natural and necessary to process in some way text and math equations together~\cite{schubotz2016semantification,yasunaga2019topiceq}. However, it is not evident how to model this in a unified framework. For instance, Yasunaga and Lafferty \cite{yasunaga2019topiceq} attempted to utilize both text and mathematical representations, but both were treated as separate components. We argue that this approach loses much crucial information, e.g., the position and the semantic dependency between text and equations.
2) Capturing semantic and structural features of math equations synchronously. Unlike text, math equations not only contain semantic features, but also structural features. For instance, equation ``$f = \frac{a}{b}$'' and ``$fb=a$'' have the same semantic features, but different structural features. However, most existing research separately considers only one of these two characteristics. 
For instance, this work \cite{yuan2016mathematical,zanibbi2016multi} only considered the structural information of equations for mathematical information retrieval tasks while other work \cite{deng2017image,yasunaga2019topiceq} treated a math equation as basic symbols and modeled them as text, which led to structural features loss.

To address these issues, we propose MathSum, a novel method that combines pointers with  multi-head attention for mathematical representation augmentation. The pointer mechanism can either copy textual tokens or math tokens from source questions in order to generate math headlines. The multi-head attention mechanism is designed to enrich the representation of each math equation separately by modeling and integrating both semantic and structural features. For evaluation, we construct two large datasets (EXEQ-300k and OFEQ-10k) which contain 290,479 and 12,548 detailed questions with corresponding math headlines from \textit{Mathematics Stack Exchange} and \textit{MathOverflow}, respectively. We compare our model with several abstractive and extractive baselines. Experimental results demonstrate that our model significantly outperforms several strong baselines on the {\task} task.

In summary, the contributions of our work are:
\begin{quote}
    \begin{itemize}
    \item an innovative {\task} task for generating a concise math headline in response to giving a detailed math question.
    \item a novel summarization model MathSum that addresses the essential issues of the {\task} task, in which the textual and mathematical information can be jointly modeled in a unified framework; while both semantic and structural features of math equations can be synchronously captured.
    \item novel math datasets
    \footnote{https://github.com/yuankepku/MathSum}. To the best of our knowledge, these are the first mathematical content/question datasets associated with headline information.
\end{itemize}
\end{quote}

\section{Related Work}

\subsection{Mathematical Equation Representation}  
Unlike text, math equations are often highly structured.
They not only contain semantic features, but also structural features. Recent work~\cite{roy2016equation,zanibbi2016multi,yuan2018formula,jiang2018mathematics} focused mainly on the structural features of math equations, and utilized tree structures to represent equations for mathematical information retrieval and mathematical word problem solving. Other work~\cite{gao2017preliminary,krstovski2018equation,yasunaga2019topiceq} instead focused mainly on the semantic features of equations. They processed an equation as a sequence of symbols in order to learn its representation. 

\subsection{Mathematical Equation Generation}
Similar to text generation, math equation generation has been widely explored. Recent work \cite{deng2017image,zhang2019improved,le2019pattern} utilized an end-to-end framework to generate equations from mathematical images, e.g., handwritten math equations. Other work~\cite{roy2016equation,wang2018mathdqn} inferred math equations for word problem solving. However, this work only supported limited types of operators (i.e., $+$, $-$, $*$, $/$).
The work~\cite{yasunaga2019topiceq} most related to ours created a model to generate equations given specific topics (e.g., electric field).
Our task~(\task), instead, aims at generating math headlines from both equations and text without clear topics. 
Thus, our \task~is quite challenging since it requires models to generate correct equations in the correct positions in the generated headlines.

\begin{table*}[t]
    \centering
    \begin{tabular}{c|cc|cc|cc|cc|cc|cc}
    \toprule
        \multirow{2}*{Datasets} &  \multicolumn{2}{c|}{avg. math num}& \multicolumn{2}{c|}{avg. text tokens} & \multicolumn{2}{c|}{avg. math tokens} &
        \multicolumn{2}{c|}{avg. sent. num} &
        \multicolumn{2}{c|}{text vocab. size}&
        \multicolumn{2}{c}{math vocab. size}\\
        & ques.&headl.&ques.&headl.&ques.&headl.&ques.&headl.&ques.&headl. &headl.&ques.\\
    \midrule
        EXEQ-300k & 6.08&1.72&60.65&7.72&12.27&9.91&4.68&1.52&84,272&21,568&1,049&663\\
        OFEQ-10k &8.56&1.41&105.92&8.61&10.04&6.84&6.53&1.40&25,733&6,721&581&393\\
    \bottomrule
    \end{tabular}
    \caption{Statistics of the EXEQ-300k and OFEQ-10k (where avg. math num = average math equation number; avg. text tokens = average textual token number; avg. math tokens = average math equation token number; avg. sent. num = average sentence number; text vocab. size = text vocabulary size; math vocab. size = math vocabulary size; ques. = detailed question (source); headl. = math headline (target)).}
    \label{tab:my_label2}
\end{table*}

\begin{table}[t]
    \centering
    \begin{tabular}{c|c|c}
    \toprule
        datasets   & question pairs & correct question pairs\\
        \midrule
        EXEQ-300k & 346,202& 290,4794\\
        OFEQ-10k &13,408 & 12,548\\
    \bottomrule
    \end{tabular}
    \caption{Statistics of two datasets (EXEQ-300k and OFEQ-10k) with respect to overall number of collected question pairs and the number of correct question pairs.}
    \label{tab:my_label1}
\end{table}


\subsection{Summarization and Headline Generation}
Summarization, a fundamental task in Natural Language Processing (NLP), can be categorized basically into extractive methods and abstractive methods.
Extractive methods~\cite{mihalcea2004textrank,nishikawa2014learning} extract sentences from the original document to form the summary. Abstractive methods~\cite{see2017get,tan2017abstractive,narayan2018don,gavrilov2019self} aim at generating the summary based on understanding the document.

We view headline generation as a special type of summarizaton, with the constraint that only a short sequence of words is generated and that it preserves the essential meaning of a math question document.
Recently, headline generation methods with end-to-end frameworks~\cite{tan2017neural,narayan2018don,zhang2018question,gavrilov2019self} achieved significant success. Math headline generation is similar to existing headline generation tasks, but still differs in several aspects.
The major difference is that a math headline consists of text and math equations which require jointly modeling and inferring text and math equations.

\begin{figure}
    \centering
    \includegraphics[scale=0.9]{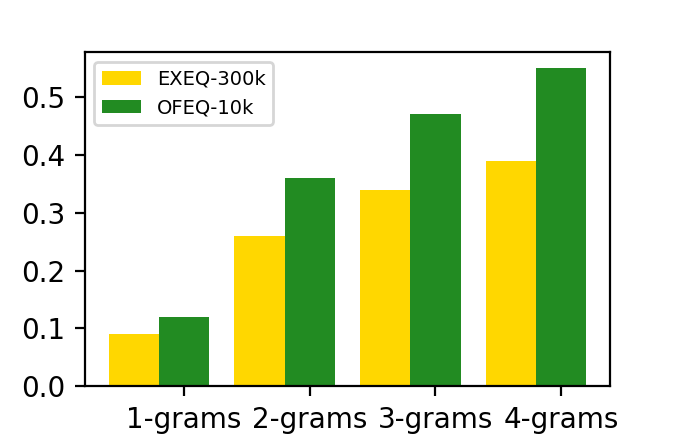}
    \caption{Proportion of novel n-grams for the gold standard math headlines in EXEQ-300k and OFEQ-10k.}
    \label{fig:his}
\end{figure}{}

\section{Task and Dataset}
\subsection{Task Definition}
Let us define the {\task} task as a summarization one. Let $\mathcal{S}=(s_0,s_1,...,s_N)$ denote the sequence of the input detailed question. $N$ is the number of tokens in the source, $s\in\{s^w,s^e\}$, $s^w$ represents the textual token (word), and $s^e$ indicates the math token\footnote{Math token is the fundamental element which can form a math equation\cite{deng2017image}}. For each input $\mathcal{S}$, there is a corresponding output math headline with $M$ tokens $\mathcal{Y}=(y_0,y_1,...,y_M)$ where $y\in\{y^w,y^e\}$ and $y^w$, $y^e$ are textual tokens and math tokens, respectively. The goal of {\task} is to generate a math headline learned from the input question, namely, $\mathcal{S}\rightarrow \mathcal{Y}$.

\subsection{Dataset}
Since this {\task} task is new, we could find no public benchmark dataset. As such, we build two real-world math datasets, EXEQ-300k (from \textit{Mathematics Stack Exchange}) and  OFEQ-10k (from \textit{MathOverflow}), for model training and evaluation. Both datasets consist of detailed questions with corresponding math headlines.

In EXEQ-300k and OFEQ-10k, each question is written in detailed math, and the corresponding headline is a human-written question summary with math equations, typically by the questioner. In \textit{Mathematics Stack Exchange} and \textit{MathOverflow}, math equations are enclosed by the ``\$\$'' symbols. We use in our datasets ``$<$m$>$'' and ``$<$/m$>$'' to replace ``\$\$'' in order to indicate the begin and end of an equation. In addition, The toolkit \texttt{Stanford CoreNLP}\footnote{https://stanfordnlp.github.io/CoreNLP/} and \LaTeX ~tokenizer in \texttt{im2mark}\footnote{https://github.com/harvardnlp/im2markup} are used to tokenize separately the text and equations in questions and headlines.

Specifically, we collect 346,202 pairs of $<$detailed questions, math headline$>$ from \textit{Mathematics Stack Exchange} and 13,408 pairs from \textit{MathOverflow}. To help with analysis and ensure quality, we remove pairs which contain math equations that cannot be tokenized by \LaTeX~tokenizer. This results in 290,479 pairs from \textit{Mathematics Stack Exchange} which form EXEQ-300k and 12,548 pairs from \textit{MathOverflow} which form OFEQ-10k. See  Table~\ref{tab:my_label2} and 
Table~\ref{tab:my_label1} for more details. In EXEQ-300k, on average there are respectively 6.08 and 1.72 math equations in the question and headlines. In contrast, OFEQ-10k contains more math equations in the question (8.56) and less in the headline (1.41).
In EXEQ-300k the questions have 60.65 textual tokens and 12.27 math tokens on average, while the headline has 7.72 textual tokens and 9.91 math tokens on average. Correspondingly, in OFEX-10k, there are on average 105.92 textual tokens and 10.04 math tokens in the question, and on average 10.04 textual tokens and 6.84 math tokens in the headline. Compared to EXEQ-300k, OFEX-10k contains more tokens (textual token and math token) in questions, and less in headlines. From Figure~\ref{fig:his}, we also see that OFEQ-10k has a higher proportion of novel n-grams than EXEQ-300k. Based on the above observations, we believe that the constructed datasets are significantly different and mutually complementary.


\begin{figure*}[t]
    \centering
    \includegraphics[scale=0.70]{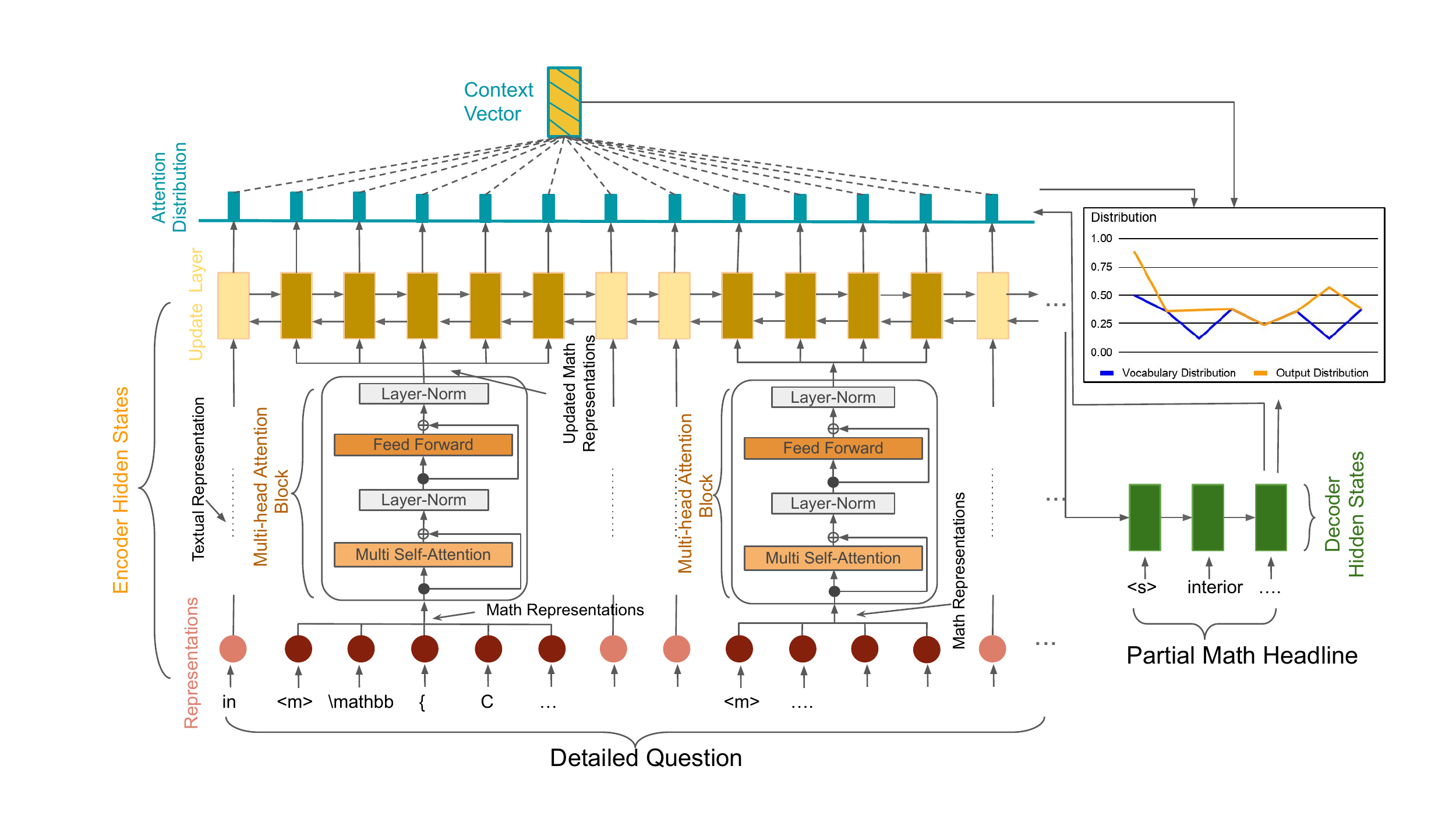}
    \caption{Architecture of MathSum. For a question, each math equation vector representation $[\mathbf{s}^e_j,...,\mathbf{s}^e_{j+m}]$ will pass through a multi-head attention block to produce a new vector representation $\mathbf{\overline{s}}^e_j$ to $\mathbf{\overline{s}}^e_{j+m}$ which updates the original representation. The updated vector representation $[\mathbf{s}^{'}_{0},...,\mathbf{s}^{'}_{N}]$ is then fed into an update layer one-by-one.}
    \label{fig:framework}
\end{figure*}
\section{Approach}

Here we describe our proposed deep model, MathSum, which we designed for the \task ~task. 
\subsection{MathSum Model}
As shown in Figure~\ref{fig:framework}, MathSum utilizes a pointer mechanism with a mutli-head attention mechanism for mathematical representation augmentation. It consists of two main components: (1) an encoder which jointly learns the representation of math equations and text, (2) a decoder which learns to generate headlines from the learned representation.

For the encoder, the crucial issue is to build effective representations for tokens in an input question. As mentioned in {\task} task, there are two different token types (i.e., textual and math) and their characteristics are intrinsically different. Math tokens not only contain the semantic features (mathematical meaning) but also the structural features (e.g., super/sub script, numerator/denominator, recursive structure). Therefore, the representation learning should vary according to the token type. In this study, we introduce a multi-head attention mechanism to enrich the representation of math tokens.

The token $s_i$ of the input question $\mathcal{S}$ is first converted into a continuous vector representation $\mathbf{s}_i$, so that the vector representation of the input is $\mathbf{S}=[\mathbf{s}_0,...,\mathbf{s}_N]$ where $N$ is the number of tokens in the input and $\mathbf{s}^w$, $\mathbf{s}^e$ are vector representation of textual and math tokens, respectively.
Then the vectors of math tokens within an equation are fed into a block with multi-head attention~\cite{vaswani2017attention} which then enriches its representation by considering both its semantic and structural features.
Please note that each equation in the input will be separately fed into the block since an equation is a fundamental unit for characterizing the semantic and structural features of a series of  math tokens.
Let $M_k=\{\mathbf{s}^e_j,...,\mathbf{s}^e_{j+m}\}$ denote the initial vector representation of the $k$-th math equation with $m$ math tokens as input. Then the multi-head attention block transforms the $\mathbf{s}^e_i$ to its enriched representation $\mathbf{\overline{s}}^e_i$.
This is calculated by 
\begin{equation}
    \mathbf{\overline{s}}^e_i = f_{\rm Multi-head}(\mathbf{s}^e_i,[\mathbf{s}^e_j,...,\mathbf{s}^e_{j+m}]), i\in\{j,..,j+m\}
\end{equation}
where $f_{\rm Multi-head}$ is the multi-head attention block.
$j$ is the beginning index of math equation $\mathcal{M}_k$ and $j+m$ is the end index.

After that, the enriched vector representation of the input is $\mathbf{S}^{'}=[\mathbf{s}^{'}_{0},...,\mathbf{s}^{'}_{N}]$ where $\mathbf{s}^{'}\in \{\mathbf{s}^{w}, \mathbf{\overline{s}}^{e}\}$ is fed into the update layer (a single-layer bidirectional LSTM) one-by-one. The hidden state $h_i$ is updated according to the previous hidden state $h_{i-1}$ and current token vector $\mathbf{s}^{'}_i$,
\begin{equation}
    h_i = f(h_{i-1}, \mathbf{s}^{'}_i)
\end{equation}
where $f$ is the dynamic function of LSTM unit and $h_i$ is the hidden state of token $\mathbf{s}^{'}$ in the step $i$.

In the decoder, we aggregate the encoder hidden states $h_0,...,h_N$ using a weighted sum that then becomes the context vector $C_t$:
\begin{equation}
    C_t = \sum_{i} \alpha_{it}h_i
\end{equation}
where

\begin{equation}
\begin{aligned}
  &  \quad\quad\quad \alpha_{t} = \rm softmax(e_t) \\
  & \rm e_{it} = \upsilon^T \rm tanh(W_{h}h_i+W_{h^{'}}h^{'}_t+b_{attn})
\end{aligned}
\end{equation}

$\upsilon$,$W_{h}$,$W_{h^{'}}$ and $b_{attn}$ are the learnable parameters.
$h^{'}_{t}$ is the hidden state of the decoder at time step $t$. The attention $\alpha$ is the distribution over the input position. 

At this point, the generated math headline may contain textual tokens or math tokens from the source which could be out-of-vocabulary. Thus, we utilize a pointer network~\cite{see2017get} to directly copy tokens from source.
Considering that the token $w$ maybe copied from the source or generated from the vocabulary, we use the copy probability $p_c$ as a soft switch to choose copied tokens from the input or generated textual tokens from the vocabulary. 
\begin{equation}
    \begin{aligned}
       & p(y_t=w|\mathcal{S},y_{<t})=p_c\sum_{i:w_i=w}\alpha_{it}+(1-p_c)f(h^{'}_t,C_t)\\
      & \quad\quad\quad\quad\quad\quad\quad p_c = f(C_t,h_t^{'},x_t)
    \end{aligned}
\end{equation}
where $f$ is non-linear function and $x_t$ is the decoder input at timestep $t$. 

Finally, the training loss at time step $t$ is defined as the negative log likehood of the target word $w_t^{*}$ where
\begin{equation}
     Loss_t = -\log p(y_t=w_t^{*}|\mathcal{S},y_{<t})
\end{equation}


\section{Experimental Setup}
\subsection{Comparison of Methods}
We compare our model with baseline methods on both the EXEQ-300k and  OFEQ-10k for the {\task} task.
Four extractive methods are implemented as baselines: \textbf{Random},  randomly selects a sentence from the input question. \textbf{Lead}, simply selects the leading sentence from the input question, while \textbf{Tail} selects the last sentence, and \textbf{TextRank}\footnote{For TextRank, we use the implementation in summanlp, https://github.com/summanlp/textrank} extracts sentences from the text according to their scores computed by an algorithm similar to PageRank. In addition, three abstractive methods\footnote{We use the implementation of OpenNMT, https://github.com/OpenNMT/OpenNMT-py} are also used to compare against MathSum. \textbf{Seq2Seq} is a sequence to sequence model based on the LSTM unit and attention mechanism~\cite{bahdanau2014neural}. \textbf{PtGen} is a pointer network which allows copying tokens from the source~\cite{see2017get}. \textbf{Transformer} is a  neural network model that is designed based on a multi-head attention mechanism~\cite{vaswani2017attention}. 

\subsection{Experiment Settings}
We randomly split EXEQ-300k into training (90\%, 261,341), validation (5\%, 14,564), and testing (5\%, 14,574) sets. In order to get enough testing samples, we split OFEQ-10k in a 80\% training (10,301), 10\% validation (1,124), and 10\% testing (1,123) proportions\footnote{For a fair comparison, all models used the same experimental data setup. For EXEQ-300k, all models are trained and tested on the same dataset. For OFEQ-10k, in order to achieve better experimental results, all models are first trained on the training set of EXEQ-300k, then fine-tuned and tested using OFEQ-10k}.

For our experiments, the dimensionality of the word embedding is 128 and the number of hidden states for LSTM units for both encoder and decoder is 512. The multi-head attention block contains 4 heads and 256-dimensional hidden states for the feed-forward part. The model is trained using AdaGrad~\cite{duchi2011adaptive} with a learning rate of 0.2, an initial accumulator value of 0.1, and a batch size of 16. Also, we set the dropout rate as 0.3. The vocabulary size of the question and headline are both 50,000. In addition, the encoder and decoder share the token representations. At test time, we decode the math headline using beam search with beam size of 3. We set the minimum length as 20 tokens on EXEQ-300k and 15 tokens on OFEQ-10k. We implement our model in PyTorch and train on a single Titan X GPU.

\section{Experimental Results}
\subsection{Quantity Performance}

\begin{table*}[t]
    \centering
    \begin{tabular}{c|ccccc|ccccc}
    \toprule
    \multirow{2}*{Models}&\multicolumn{5}{c|}{EXEQ-300k}&\multicolumn{5}{c}{OFEQ-10k}\\
           & R1 & R2 & RL&BLEU-4&METEOR& R1 & R2 & RL&BLEU-4&METEOR\\
        \midrule
        Random &31.56&21.35&28.99&24.32&23.40 &22.95&11.48&19.85&13.19&18.00\\
        Tail & 22.55&14.69&20.76&22.23&23.78&15.46&7.03&13.36&11.13&11.68\\
        Lead & 42.23&31.30&39.29&29.89&31.61&27.68&14.92&24.07&14.56&20.99\\
        TextRank &42.19&30.85&38.99&28.29&31.78&29.66&16.41&25.59&14.20&23.71 \\
        \midrule
        Seq2Seq &52.14&38.33&49.00&42.20&30.65&38.64&23.42&35.24&27.67&25.27 \\
        PtGen &53.26&39.92&50.09&44.10&31.76&40.27&25.30&36.51&28.07&25.90\\
        Transformer& 54.49&40.57&50.90&45.79&32.92&40.54&24.36&36.39&28.82&25.89\\
        \midrule
        MathSum&\textbf{57.53}&\textbf{45.62}&\textbf{54.81}&\textbf{52.00}&\textbf{37.47}&\textbf{42.44}&\textbf{28.15}&\textbf{38.99}&\textbf{29.44}&\textbf{26.84}\\
    \bottomrule
    \end{tabular}
    \caption{Comparison of different models on the EXEQ-300k and OFEQ-10 test sets for \textbf{F1} scores of R1 (ROUGE-1), R2 (ROUGE-2), RL (ROUGE-L), BLEU-4, and METEOR.} 
    \label{tab:my_label3}
\end{table*}

\begin{table*}[]
    \centering
    \begin{tabular}{c|ccc|ccc}
    \toprule
    \multirow{2}*{Models} &\multicolumn{3}{c|}{EXEQ-300k}&\multicolumn{3}{c}{OFEQ-10k}\\
         & Edit Distance(m)&Edit Distance(s)&Exact Match&Edit Distance(m)&Edit Distance(s)&Exact Match \\
        \midrule
         Random& 8.76&21.84&9.29&7.20&17.73&5.60\\
         Tail&9.42&20.89&6.65&7.30&14.45&3.47\\
         Lead&7.47&20.27&12.39&6.58&17.75&6.70\\
         TextRank&7.68&21.36&12.68&6.75&20.27&7.71\\
         \midrule
         Seq2Seq&6.68&13.57&13.26&8.69&16.78&8.68\\
         PtGen&6.59&13.43&13.60&8.06&15.56&8.56\\
         Transformer&6.32&13.23&13.94&\textbf{5.56}&\textbf{10.51}&8.41\\
         \midrule
         MathSum&\textbf{5.82}&\textbf{12.07}&\textbf{15.21}&5.71&10.76&\textbf{8.98}\\
         \bottomrule
    \end{tabular}
    \caption{Comparison of different models on the EXEQ-300k and OFEQ-10k test sets according to math evaluation metrics. Edit Distance(m) and Edit Distance(s) evaluate those that are dissimilar (the smaller the better). Exact Match is the number of math tokens accurately generated in math headlines (the larger the better).}
    \label{tab:math_result}
\end{table*}

\subsubsection{Metrics}
Here we use three standard metrics: ROUGE~\cite{lin2004rouge}, BLEU~\cite{papineni2002bleu} and METEOR~\cite{denkowski2014meteor} for evaluation. The ROUGE metric measures the summary quality by counting the overlapping units (e.g., n-gram) between the generated summary and reference summaries. We report the F1 scores for R1 (ROUGE-1), R2 (ROUGE-2), and RL (ROUGE-L). The BLEU score is a widely used as an accuracy measure for machine translation and computes the n-gram precision of a candidate sequence to the reference. METEOR is recall-oriented and evaluates translation hypotheses by aligning them to reference translations and calculating sentence-level similarity scores. The BLEU and METEOR scores are calculated by using \texttt{nlg-eval}\footnote{https://github.com/Maluuba/nlg-eval} package, and ROUGE scores are based on \texttt{rouge-baselines}\footnote{https://github.com/sebastianGehrmann/rouge-baselines} package.

We use the edit distance and exact match to check the similarity of the generated equations compared with the gold standard equations in the math headlines. These two metrics are widely used for the evaluation of equation generation~\cite{deng2017image,wu2018image}. Edit distance quantifies how dissimilar two strings are by counting the minimum number of operations required to transform one string into the other. Based on $N$ samples in the test set, we use two types of edit distance. One is Edit Distance(m) which is math-level dissimilar score and is defined as $Edit Distance(m)=\sum^{N}_{i=0}\frac{minMd_i}{\max(|P_i|, |G_i|)}$, where $minMd$ is the minimum edit distance between equations in the generated headline and the gold standard headline, $|P_i|$ and $|G_i|$ are the number of equations in the $i$-th generated headline and gold headline. The other Edit Distance(s) is the sentence-level dissimilar score, and is formulated as $Edit Distance(s)=\frac{\sum^{N}_{i=0}minMd_i}{N}$.  Exact Match checks the exact match accuracy between the gold standard math tokens and generated math tokens and is calculated as $Exact Match=\frac{\sum^{N}_{i=0}(PM_i \& GM_i)}{N}$, where $PM_i$ and $GM_i$ are the sets of math tokens in the $i$-th generated headline and gold standard headline.

\subsubsection{Results}
Comparisons of models can be found in Table~\ref{tab:my_label3}. All models perform better on EXEQ-300k than OFEQ-10k. A possible explanation is that the EXEQ-300k contains a lower proportion of novel n-grams in its gold standard math headlines (illustrated in Figure~\ref{fig:his}). 
For extractive models, we find that Lead obtains a good performance on EXEQ-300k, while TextRank performs well on OFEX-10k. Since OFEX-10k contains more sentences for each question, TextRank is more likely to pick out the accurate sentence. Unsurprisingly, abstractive models perform better than extractive models on both datasets. Compared to ordinary Seq2Seq, PtGen gets  better performance, since it uses a copying strategy to directly copy tokens from the source question. The transformer can outperform PtGen, which implies that by utilizing multi-head attention mechanism, we obtain a better learning of representation. MathSum significantly outperforms other models for all evaluation metrics on both datasets. Thus, MathSum initially addresses some of the challenges of {\task} task and generates satisfactory headlines for questions.

\begin{figure*}
    \centering
    \includegraphics[scale=0.95]{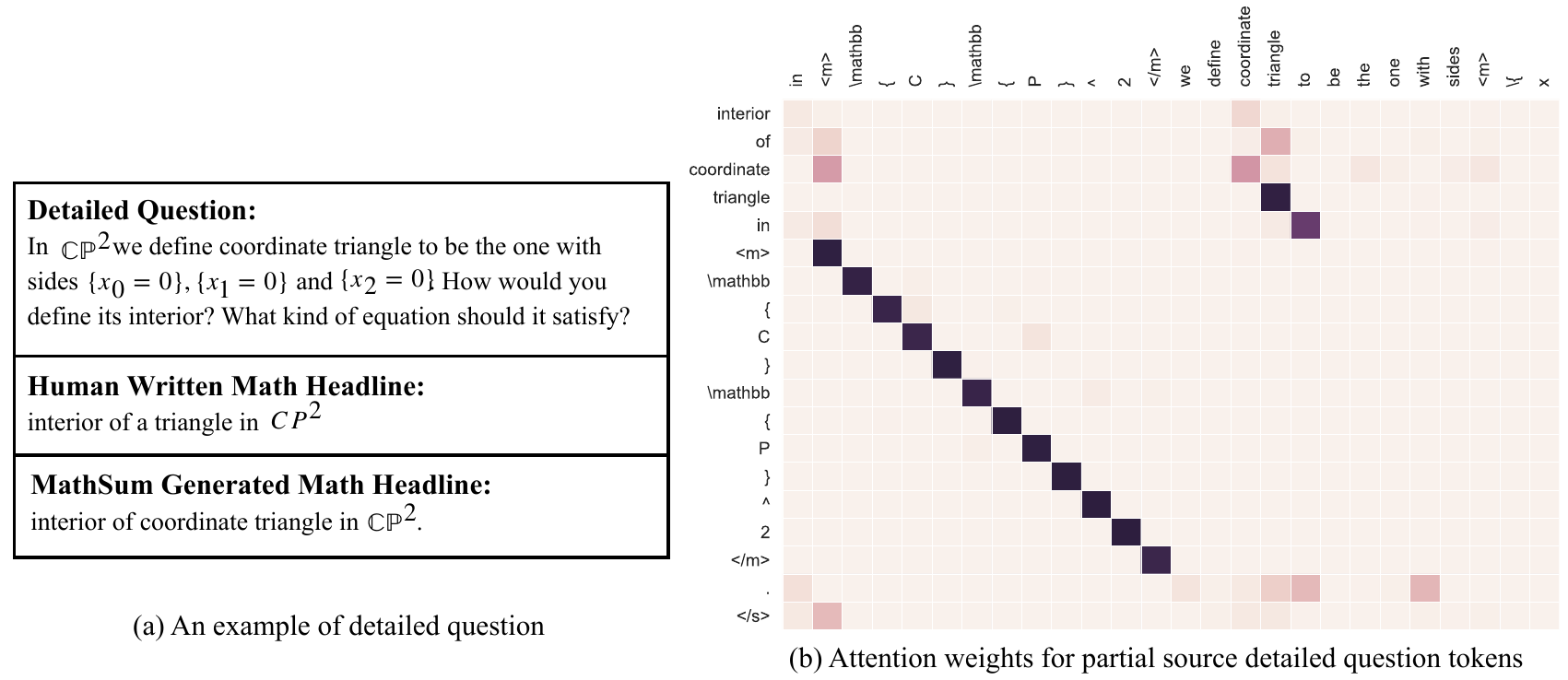}
    \caption{Heatmap of attention weights for source detailed questions. MathSum learns to align key textual tokens and math tokens with the corresponding tokens in the source question.}
    \label{fig:attention}
\end{figure*}{}

In addition, we also evaluate the gap between the generated headlines and human-written headlines. The Edit Distance(m), Edit Distance(s) and Exact Match scores for different models using EXEQ-300k and OFEQ-10k are shown in Table~\ref{tab:math_result}.
The results show that extractive models perform worse, if we use the metric Edit Distance(s) instead of Edit Distance(m) for evaluation. Since extractive models directly select sentences from source questions, some selected sentences may not contain math equations. For abstractive baselines,  
the Transformer obtains the best performance. This observation reinforces the claim that a mutli-head attention mechanism can construct a better representation for math equations. On EXEQ-300k, our model, MathSum, achieves the best performance on all metrics. On OFEX-10k, MathSum gets the best performance for Exact Match and second best performance (slightly weaker than Transformer) for Edit Distance(m) and Edit Distance(s). A possible reason is that in OFEX-10k, the lengths of math equations in source questions are usually long, while the ones in headlines are often short. Compared to the Transformer, the copying mechanism could cause MathSum to copy long equations from the source questions, which may result in a slight decreased performance for Edit Distance(m) and Edit Distance(s) metrics.

\subsection{Quality Analysis}
\subsubsection{Jointly modeling quality}

The heatmap in Figure~\ref{fig:attention} visualizes the attention weights from MathSum.
Figure~\ref{fig:attention}(a) compares the source detailed question with its human-written math headline and the generated math headline from MathSum. As Figure~\ref{fig:attention} shows, there are both textual tokens and math tokens in the generated headline. Note that both math tokens and textual tokens can be effectively aligned to their corresponding tokens in the source. For instance, the textual tokens ``coordinate", ``triangle" and the math tokens ``$P$", ``$C$" are both all successfully aligned.
\subsubsection{Case study}
To gain an insightful understanding regarding the generation quality of our method, we present three typical examples in Table~\ref{tab:examples}. The first two are selected from EXEQ-300k\footnote{https://math.stackexchange.com/questions/2431575}$^,$\footnote{ https://math.stackexchange.com/questions/752067} and the last one is selected from OFEQ-10k\footnote{https://mathoverflow.net/questions/291434}. From the examples, we see that the generated headlines and the human-written headlines have comparability and similarity. Generally, the generated headlines are coherent, grammatical, and informative. 
We also observe that, it is important to locate the main equations for {\task} task. If the generation method emphasizes a subordinate equation, it will generate an unsatisfactory headline, such as the second example in Table~\ref{tab:examples}. 

\section{Conclusions and Future Work}
Here we define and explore the novel {\task} task of automatic headline generation for online math questions using a new deep model, MathSum. Two new datasets (EXEQ-300k and OFEQ-10k)
are constructed for algorithm training and testing and are made available. Our experimental results demonstrate that our model can often generate useful math headlines and significantly outperform a series of state-of-the-art models.
Future work could focus on enriched representations of math equations for mathematical information retrieval and other math-related research. 

\begin{table}[t]
\footnotesize
    \begin{tabular}{|m{2.4cm}<{\centering}|m{5.1cm}<{\centering}|}
    \hline
    \multicolumn{2}{|c|}{\textbf{Examples}}\\
        \hline
         Partial Math Detailed
         Question (EXEQ-300k) & So I am asked to find the inverse elements of this set $\mathbb{Z}[i] = \{ a + ib | a,b \in \mathbb{Z} \}$ (I know that this is the set of Gaussian integers). I was pretty much do...\\
         \hline
         Human-Written & finding the inverse elements of $\mathbb{Z}[i] = \{ a + ib | a,b \in \mathbb{Z} \}$ \\
         \hline
         MathSum & finding the inverse elements of $\mathbb{Z}[i] = \{ a + ib | a,b \in \mathbb{Z} \}$\\
         \hline
         Partial Math Detailed
         Question (EXEQ-300k) &Suppose that the function $\psi:\mathbb{R}^2 \to \mathbb{R}$ is continuously differentiable. Define the function $g:\mathbb{R}^2 \to \mathbb{R}$ by...\\ 
         \hline
         Human-Written&using the chain rule in $\mathbb{R}^n$\\
         \hline
         MathSum &find $\frac{\partial g}{\partial s}(s,t)$\\
         \hline
         \hline
         Partial Math Detailed Question (OFEQ-10k) & In the paper of Herbert Clemens Curves on generic hypersurfaces the author shows that for a generic hypersurface $V$ of ${\mathbb P}^n$ of sufficiently high degree there is no rational...  \\
         \hline
         Human-Written & rational curves in ${\mathbb P}^n$ and immersion \\
         \hline
         MathSum & rational curves in ${\mathbb P}^n$ \\
         \hline
    \end{tabular}
    \caption{Examples of generated math headlines given detailed questions.} 
    \label{tab:examples}
\end{table}

\section{Acknowledgments}
This work is partially supported by China Scholarship Council and projects of National Natural Science Foundation of China (No. 61876003 and 61573028), Guangdong Basic and Applied Basic Research Foundation (2019A1515010837) and
Fundamental Research Funds for the Central Universities (18lgpy62),
and the National Science Foundation.
\bibliographystyle{aaai}
\bibliography{ref}

\end{document}